\documentclass[10pt,twocolumn,letterpaper]{article}

\usepackage{cvpr}
\usepackage{multirow}
\usepackage{pifont}
\usepackage[dvipsnames]{xcolor}

\definecolor{cvprblue}{rgb}{0.21,0.49,0.74}
\usepackage[pagebackref,breaklinks,colorlinks,citecolor=cvprblue]{hyperref}

\def\paperID{17} 
\def\confName{CVPR}
\def\confYear{2024}

\newcommand{\parsection}[1]{\textbf{#1} }

\title{LLM-Seg: Bridging Image Segmentation and Large Language Model Reasoning}

\author{Junchi Wang\\
ETH Zurich\\
\and
Lei Ke\\
ETH Zurich\\
}

\begin{document}
\maketitle
\begin{abstract}
Understanding human instructions to identify the target objects is vital for perception systems. 
In recent years, the advancements of Large Language Models (LLMs) have introduced new possibilities for image segmentation. In this work, we delve into reasoning segmentation, a novel task that enables segmentation system to reason and interpret implicit user intention via large language model reasoning and then segment the corresponding target. Our work on reasoning segmentation contributes on both the methodological design and dataset labeling. For the model, we propose a new framework named LLM-Seg. LLM-Seg effectively connects the current foundational Segmentation Anything Model and the LLM by mask proposals selection. For the dataset, we propose an automatic data generation pipeline and construct a new reasoning segmentation dataset named LLM-Seg40K. Experiments demonstrate that our LLM-Seg exhibits competitive performance compared with existing methods. Furthermore, our proposed pipeline can efficiently produce high-quality reasoning segmentation datasets. The LLM-Seg40K dataset, developed through this pipeline, serves as a new benchmark for training and evaluating various reasoning segmentation approaches. Our code, models and dataset are at \url{https://github.com/wangjunchi/LLMSeg}.
\end{abstract}    
\section{Introduction}
\label{sec:intro}

\begin{figure}[!htbp]
	\centering
	\includegraphics[width=\linewidth]{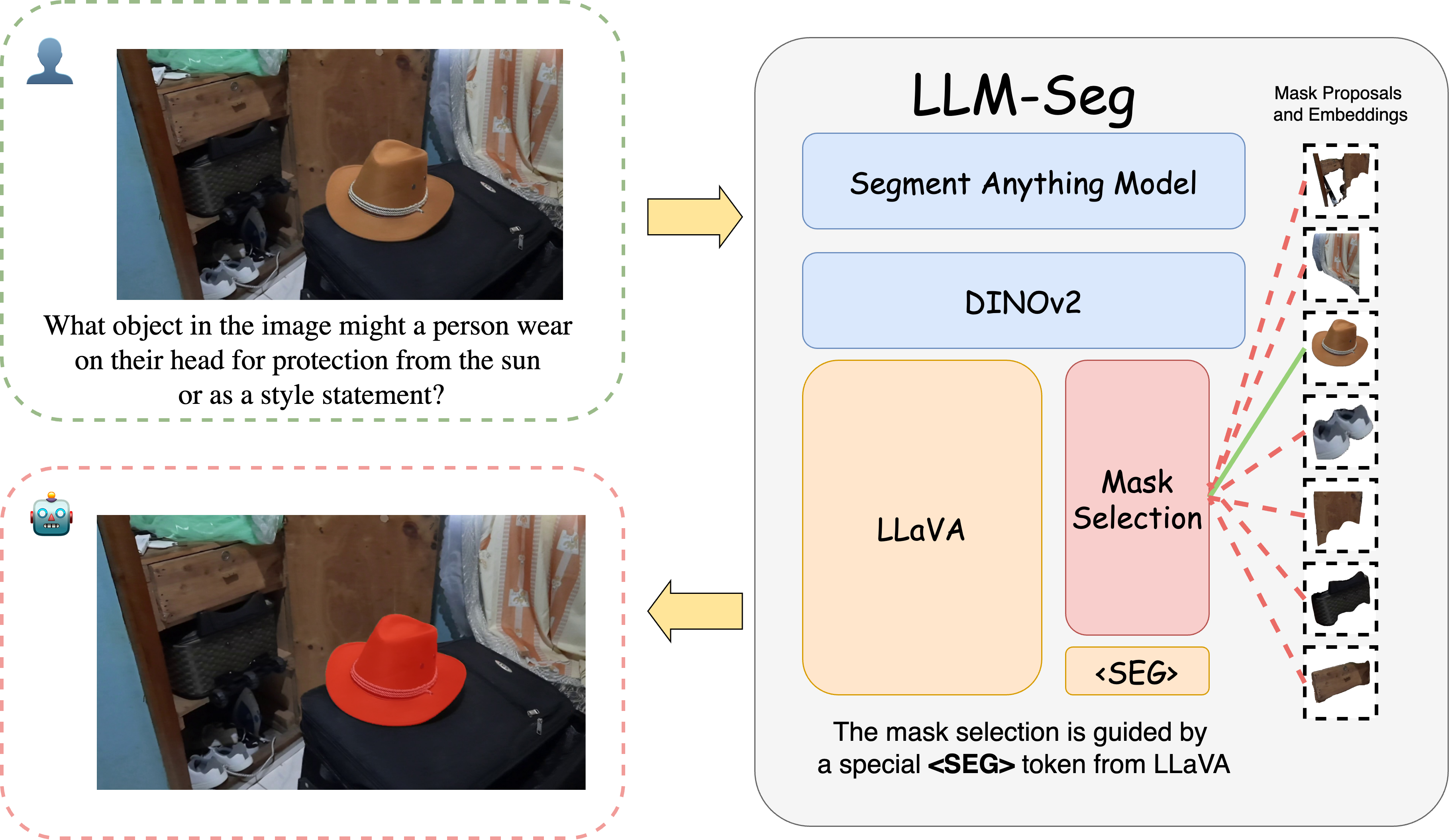}
	\caption{Our LLM-Seg model integrates multiple foundation models including LLaVA \cite{liu2023visual}, Segment Anything Model \cite{kirillov2023SAM}, and DINOv2 \cite{oquab2023dinov2}. The Segment Anything Model and DINOv2 generate mask proposals and embeddings. LLaVA is responsible for perceiving the input image and question, and it outputs a special $<\text{SEG}>$ token to guide the mask selection module.}
	\label{fig:teaser}
\end{figure}

Traditional image segmentation requires explicit target objects or pre-define categories for pixel-level classification. 
However, a broad spectrum of open-world applications require the segmentation system to understand and interact with more complicated human instructions, such as home robots \cite{li2022robotic_vision}, self-driving \cite{feng2020deep_driving}, and augmented reality \cite{ko2020novel_ar}.  

In recent years, the development of vision-language pre-trained models like CLIP \cite{radford2021CLIP}, has shifted image segmentation from close-set segmentation, where targets must belong to a predefined set of categories, to open-vocabulary segmentation, enabling adaptation to unseen categories. By unifying text and image into the same feature spaces, these models significantly reduce the cost of training across new datasets, thereby broadening the applicability of image segmentation tasks. However, the scope of open-vocabulary segmentation remains constrained to the vocabulary or phrase level, lacking the ability to understand long and complex text prompts. 

The recent success of Large Language Models (LLM) \cite{almazrouei2023falcon, bai2022constitutional, brown2020GPT3, chowdhery2023palm, touvron2023llama} brings new possibilities to how the segmentation target is defined. The state-of-the-art LLM models such as ChatGPT and LLama \cite{touvron2023llama} show incredible reasoning ability and can answer complex questions. To transfer the reasoning ability of LLM into image segmentation, \cite{lai2023lisa} proposes Reasoning Segmentation. This new image segmentation task requires the model to segment the target based on a question asked by a human. Reasoning Segmentation is a more challenging task because it requires strong reasoning ability to find out the segmentation target, as well as its unclear granularity of the target. However, the application of Reasoning Segmentation is more extensive and dynamic. 

In this work, we propose LLM-Seg: a two-stage method that combines vision language model (VLM) and vision foundation models. We define our method as two-stage because it decouples processes of prompt understanding and image segmentation. In our LLM-Seg, a frozen Segment Anything Model is utilized to generate a series of mask proposals. We adopt the LLaVA, which is one of the state-of-the-art VLMs, to receive the prompt from the user and select from the mask proposals. We prefer the two-stage method over an end-to-end method such as \cite{lai2023lisa} for the following reasons. Firstly, current vision foundation models, exemplified by the Segment Anything Model (SAM) \cite{kirillov2023SAM}, have been trained using extensive datasets and substantial GPU resources. Consequently, fine-tuning such models with limited data could harm their performance and generality. Secondly, the two-stage method is more flexible in the choice of components. Finally, decoupling the reasoning part and segmentation could simplify the learning target, resulting in faster convergence during training.

One challenge for the research of reasoning segmentation is the scarcity of datasets. In response to this challenge, we additionally propose a new method that leverages ChatGPT-4 to process existing semantic segmentation datasets and automatically generate a dataset tailored for reasoning segmentation. Utilizing the ChatGPT-4 API enables the generation of high-quality questions for training and evaluation. Furthermore, reusing existing semantic segmentation datasets alleviates the burden of annotating new images, and ensures high-quality ground truth data for the dataset.

The contribution of our research are delineated as follows:
\begin{itemize}
    \item We introduce LLM-Seg, a novel two-stage method tailored for the task of reasoning segmentation. 

    \item We develop a low-cost and efficient data generation pipeline for reasoning segmentation, capitalizing on the advanced capabilities of ChatGPT-4.

    \item We propose the LLM-Seg40K, a large-scale reasoning segmentation dataset for model training and evaluation.

\end{itemize}

\section{Related Work}
\label{sec:formatting}

\parsection{Vision Language Models}
The success of Large Language Models in recent years has revolutionized the field of natural language processing. LLMs such as GPT-3 \cite{brown2020GPT3}, Chinchilla \cite{hoffmann2022chinchilla}, and LLama \cite{touvron2023llama} have shown powerful zero-shot performance in different language tasks and become the base model of many other LLMs. Based on powerful base language models, many methods try to add visual information to the input domain. 
Existing VLMs demonstrate diverse strategies for integrating visual and textual information. Flamingo  \cite{alayrac2022flamingo} leverages a pretrained Normalizer-Free ResNet to extract image features and a perceiver resampler module for visual token sampling. To condition the output to the visual tokens, gated cross-attention layers are added between the layers of the frozen base model. BLIP-2  \cite{li2023blip} introduces cost-effective training via a lightweight Querying-Transformer  (Q-Former) Module, focusing on bridging the gap between image encoders and language models. LLaVA \cite{liu2023visual} offers a simplified structure through Visual Instruction Tuning. Given enough visual instruction data, LLaVA shows that a simple linear projection layer can achieve state-of-the-art performance in general visual-language tasks. Although VLMs show impressive ability in visual understanding, the output of current VLMs is limited to text, making it hard to adopt VLMs to tasks like image segmentation.

\parsection{Vision Foundation Models for Image Segmentation}
Similar to the fast development of Large Language Models in the NLP community, vision foundation models \cite{kirillov2023SAM, oquab2023dinov2, radford2021CLIP, xiao2023florence2, yuan2021florence} for various computer vision tasks have marked a significant technological advancement. One of the most notable strengths of these foundation models is their robust zero-shot learning abilities. This feature enables them to be seamlessly adapted to the image segmentation task, demonstrating remarkable efficacy without the need for expensive and time-consuming fine-tuning.

One of the most popular vision foundation models for image segmentation is CLIP \cite{radford2021CLIP}. CLIP demonstrates its impressive ability to extract semantic visual features and remarkable zero-shot performance in image classification tasks. Based on the pretrained CLIP model, image segmentation could benefit from CLIP's zero-shot ability. Early methods try to push the contrastive learning of CLIP from the image level to either patch level \cite{liang2023ovseg} or pixel level \cite{wang2022cris}. However, fine-tuning the CLIP parameters may harm the zero-shot ability because of the small scale of data for fine-tuning. More recent methods focus on building a segmentation model upon a frozen CLIP model such as FC-CLIP \cite{yu2023fcclip} and CLIPSeg \cite{luddecke2022clipseg}. Similar to CLIP, DINOv2 \cite{oquab2023dinov2} is another vision foundation model that can generate all-purpose image features. Experiments show that the image features from DINO-V2 give better performance than CLIP features in different vision tasks \cite{oquab2023dinov2}. 

In addition to CLIP and DINOv2, Segment Anything Model (SAM)~\cite{kirillov2023SAM} and its variants~\cite{sam_hq,mobile_sam} are another groundbreaking vision foundation models, particularly in image segmentation. SAM defines a new promptable segmentation task that unifies all the objects inside images regardless of their category, making the training dataset highly scalable and improving the zero-shot ability in image segmentation. 
However, the introduced vision foundations focus more on the zoro-shot ability of traditional vision tasks. Existing research in applying these models to multimodal tasks such as reasoning segmentation is limited. 

\parsection{Methods for Reasoning Segmentation}
In this section, we introduce some existing methods that can be applied to reasoning segmentation. Previous models \cite{liu2023gres, devlin2018bert, zou2023segment} for the referring segmentation can be directly the reasoning segmentation. Most of these models only apply a text encoder like BERT \cite{devlin2018bert} to process the text and lack complex reasoning ability. Therefore, referring to segmentation models can only be a weak baseline for the novel reasoning segmentation task.

To make use of the powerful reasoning ability of vision language models, LISA \cite{lai2023lisa} connects LLaVA with SAM and trains the model end-to-end. Specifically, a new embedding prompt is added to the mask decoder of the SAM to guide the segmentation. The vision language model is adopted to generate the embedding prompt based on the image and question. However, fine-tuning the mask decoder of SAM will significantly harm the segmentation quality, resulting in holes and coarse boundaries of the segmentation mask.

Different from end-to-end methods such as LISA, reasoning segmentation could also be solved by decoupling it into the reasoning and segmentation stages. Each stage could be completed with state-of-the-art models. For example, we could first prompt the question to any vision language model and require the output to be a simple phrase. The output phrase could be used as input for an open-vocabulary segmentation model to produce the segmentation mask. Besides, vision language models such as Shkira \cite{chen2023shikra} and CogVLM \cite{wang2023cogvlm} have the visual grounding ability and can give out the bounding box of the target object in text form. The output bounding box can be used as the box prompt of SAM to generate the segmentation mask.

\section{Method}

\subsection{Model Structure}
Our proposed LLM-Seg is composed of 4 parts: the pretrained LLaVA-7B model, the Segment Anything Model, the DINOv2 model, and the mask selection module. The vision language model is responsible for understanding the content of the input image and question. Inspired by LISA \cite{lai2023lisa}, we add a special  $<\text{SEG}>$ token to the vocabulary table of LLaVA. The $<\text{SEG}>$ token is updated during the forward pass, containing information about the segmentation target. The input of the mask proposal generator is image only, and the outputs are multiple binary masks that indicate elements in the image. The mask selection module selects from the mask proposals based on the $<\text{SEG}>$ token. The final segmentation of the model output is the combination of selected mask proposals.

\begin{figure*}[!htbp]
	\centering
	\includegraphics[width=\linewidth]{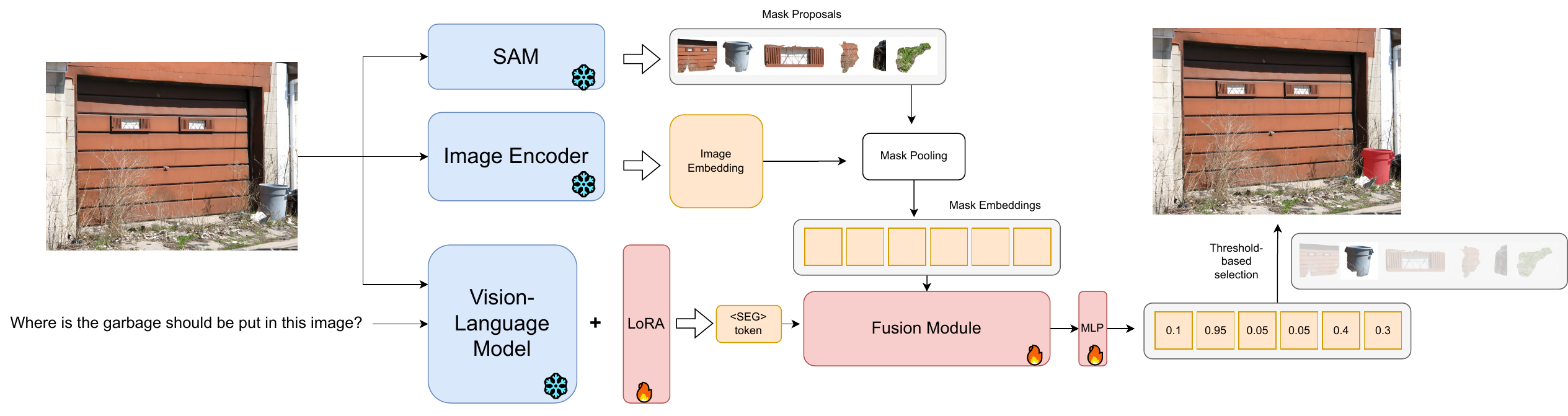}
	\caption{Model Structure of our LLM-Seg. The input image will be processed by three different modules. The SAM is responsible for generating binary mask proposals using its Everything Mode. The image encoder extracts features from the image. The vision language model processes the image together with the input queries and uses a special $<\text{SEG}>$ token to represent the result. The mask embedding will be extracted for each mask proposal using mask pooling. Using the information from mask embeddings and $<\text{SEG}>$ token, the fusion model and an MLP layer will predict a score for each mask proposal. Finally, a simple threshold-based selection is used to pick mask proposals as the final prediction.}
	\label{fig:full_structure}
\end{figure*}

The diagram of our model structure is shown in Figure \ref{fig:full_structure}. Inside the model, only the mask selection module contains full trainable parameters. The SAM and DINOv2 are completely frozen. The vision language model is optimized by LoRA \cite{hu2021lora}. In this way, we successfully control the size of trainable parameters and make sure our model can be trained using a single RTX 3090 GPU. 

\subsection{Mask Proposals}
In LLM-Seg, we choose Segment Anything Model (SAM) as the mask proposal generator and use its Everything Mode to produce binary masks. The Everything Mode of SAM is implemented based on the point prompt. Given an image of shape $H*W$, $32*32$ points are sampled uniformly within the image. Each point is an independent prompt and is responsible for generating one mask. For each mask, SAM also gives an IoU prediction. Low-IoU mask will be filtered out and duplicate masks will be removed with Non-Maximum Suppression. Although $32*32=1024$ point prompts exist for one image, the heavy image encoder only runs once, and the lightweight mask decoder can process the prompts in batches. The whole process takes about 3.5 seconds for one image using a single RTX-3090 GPU.

After generating a series of mask proposals, we apply mask pooling to convert the binary mask proposals into mask embeddings. The image features for the mask pooling come from a separate DINOv2 model.

\subsection{Mask Selection}

\begin{figure}[!htbp]
	\centering
	\includegraphics[width=\linewidth]{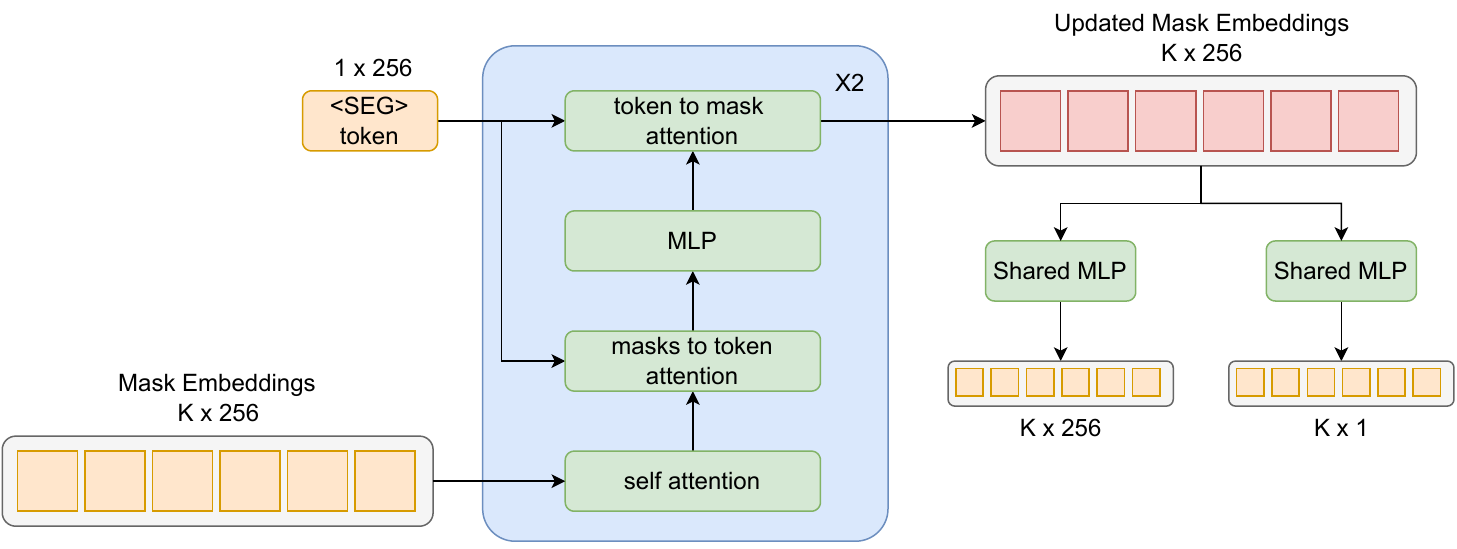}
	\caption{Diagram of the fusion module and two selection heads. The input of the fusion module is $K$ mask embeddings and $1$ $<\text{SEG}>$ token. After the fusion module, the updated mask embeddings are processed by two separate MLP layers to predict different targets. The target with $256$ dimension is used to compute the loss of the IoU head, while the target with $1$ dimension is used for IoP regression.}
	\label{fig:fusion_module}
\end{figure}

The mask selection module is composed of one mask-text fusion module and two selection heads. Similar to SAM's mask decoder, the fusion module applies self-attention and cross-attention mechanisms to align the mask embeddings and  $<\text{SEG}>$ token. However, our fusion module is more computing efficient due to the small number of mask embeddings. The diagram of the fusion module is shown in Figure \ref{fig:fusion_module}.

The two selection heads are named IoU head and IoP head. For the IoU head, we first simplify the mask selection process by choosing the mask proposal with the highest IoU with ground truth. Specifically, the selected mask proposal should have the highest similarity with $<\text{SEG}>$ token. The IoU head contains a shared MLP to map all the mask embeddings to the same embedding space as $<\text{SEG}>$ token. Then the dot product is performed to compute the similarity. Because we only care about the mask embedding with the highest similarity, a small temperature factor $\tau$ is used to normalize the similarities. We apply KL-Divergence to supervise the training, as shown in Equation \ref{eq:loss_iou}. $\mathbf{P}$ is the vector of all mask embeddings, $\mathbf{s}$ is the embedding of $<\text{SEG}>$ token, and $\mathbf{I}$ is the ground truth IoU of all mask proposals.

\begin{equation} \label{eq:loss_iou}
    \mathcal{L}_{\text{iou}} = \textbf{KL}\left( \text{Softmax}(\mathbf{P} \cdot \mathbf{s}^\top / \tau), \text{Softmax}(\mathbf{I} / \tau) \right)
\end{equation}

However, selecting only one mask proposal will not give the best performance in the multi-instance scenario. To improve the performance, an IoP head is added to select multiple masks. The selection criterion is based on the Intersection over Prediction (IoP) which can also be treated as the pixel accuracy of each mask proposal. The IoP is computed as Equation \ref{eq:iop}. $P$ is the predicted masks and $G$ is the ground truth mask.

\begin{equation} \label{eq:iop}
    \text{IoP} = \frac{|G \cap P|}{|P|}
\end{equation}

We use a shared MLP to directly regress the mask embeddings to the corresponding IoP. The selection then can be based on a threshold of the predicted IoPs. Due to the heavy imbalance of IoPs, we apply a weighted L2 loss to supervise the training as shown in Equation \ref{eq:loss_iop}. $K$ is the number of mask proposals for an image, $p_k$ is the predicted IoP for $k_{th}$ mask proposal and $g_k$ is the ground truth IoP.

\begin{equation} \label{eq:loss_iop}
\mathcal{L}_\text{iop}= \frac{1}{K} \sum_{k=1}^{K} \frac{1}{e^{(1 - g_k)}} \cdot \text{MSE}(p_k, g_k)
\end{equation}

\subsection{Model Training}
In the training process, only our proposed mask selection modules will be fully updated. We apply LoRA to fine-tune the LLaVA-7B model and freeze all parameters of SAM and DINOv2. The training loss is the combination of the losses from the IoU selection head and the IoP selection head as shown in Equation \ref{eq:loss_all}, $\lambda_{iou}$ and $\lambda_{iop}$ are the weight for two losses.

\begin{equation} \label{eq:loss_all}
    \mathcal{L} = \lambda_{iou} \mathcal{L}_{iou} + \lambda_{iop} \mathcal{L}_{iop}
\end{equation}

\section{LLM-Seg40K}
Reasoning segmentation is a novel task and research on it is still limited. Existing dataset \cite{lai2023lisa} only contains hundreds of images, which is quite small compared to classic segmentation datasets \cite{gupta2019lvis, lin2014mscoco, neuhold2017mapillary, zhou2019ade20k}. Creating a large-scale dataset is vital for the research of reasoning segmentation. This section demonstrates a novel data generation pipeline and new reasoning segmentation dataset, introduced as LLM-Seg40K.

\subsection{Dataset Definition}
We first define the format of samples in our LLM-Seg40K. Each image, represented as $x_{img}$, is associated with multiple text and segmentation pairs $\{x_{text}, y_{seg }\}$. For each pair of $\{x_{text}, y_{seg }\}$, $x_{text}$ is a question targeting one or more objects in the image, and $y_{seg}$  is a binary mask corresponding to the answer to $x_{text}$. The question should require basic knowledge and reasoning to identify the target within the image correctly. For example, instead of asking "\texttt{Where is the headphone in the image?}", the question in LLM-Seg40K should be "\texttt{Among all the electronic devices shown in the image, which object could a person use to listen to music without disturbing others?}". 

\subsection{Data Generation Pipeline}

\subsubsection{Data Sources}
Manually annotating questions and segmentation is time-consuming and expensive. Therefore, we reuse the existing image segmentation from LVIS \cite{gupta2019lvis} and EgoObjects \cite{zhu2023egoobjects} to construct our new dataset. Combining high-quality photographic images (from LVIS) and egocentric images can improve the diversity of our dataset and better reflect real applications such as robotics and virtual assistance.

For the LVIS dataset, we notice a dilemma between the diversity and complexity of the images. Images with many segmentation categories tend to concentrate on indoor and street scenes. To balance between diversity and complexity, we sample images from the two simple and complex images differently. In our definition, simple images should have at least 2 but fewer than 6 different categories and complex images should have more than 5 categories. We randomly sample 6000 images from simple images and 4000 images from complex images.

For the EgoObjects dataset, We randomly sample 3000 images among images containing more than 2 categories. In addition, We apply SAM to convert the original bounding box annotation to binary segmentation for sampled images.

\subsubsection{Prompt Engineering of GPT-4}
In this work, we utilize the text-only ChatGPT-4 to generate high-quality questions for reasoning segmentation. To help ChatGPT-4 understand the image, The first step is to convert the images to text. Our pipeline applies the LLaVA-v1.5-13B model \cite{liu2023improved} to generate the image descriptions. We query LLaVA with the prompt: ``\texttt{Please describe the content in this image within 10 sentences.}". All images are processed independently with single-turn conversation.

\begin{figure}[t]
	\centering
	\vspace{-0.15in}
\includegraphics[width=1.1\linewidth]
{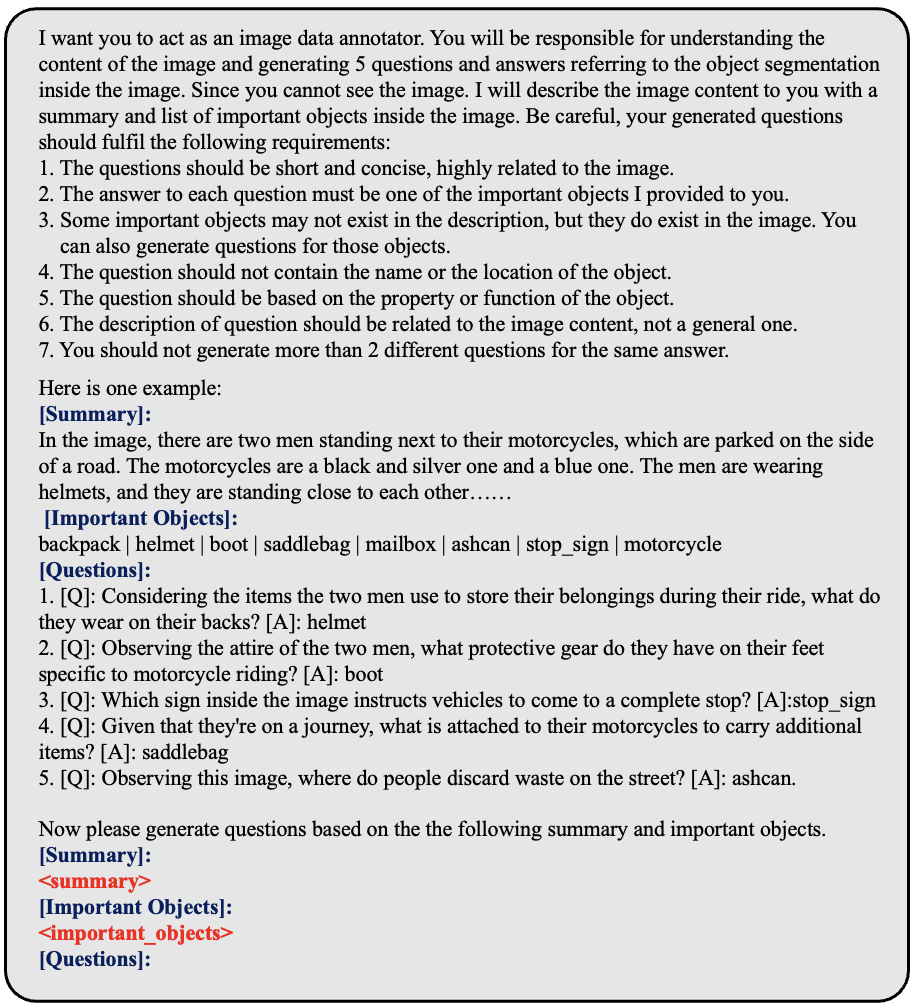}
\vspace{-0.15in}
	\caption{The complete prompt template used to prompt the ChatGPT-4. The example part is fixed for all the queries and we only replace the $<\text{summary}>$ and $<\text{important\_objects}>$ which are highlighted in red.}
	\label{fig:gpt_prompt}
	\vspace{-0.1in}
\end{figure}

After converting images to text, we can make use of the text-only ChatGPT-4 to generate high-quality questions for reasoning segmentation. We carefully design the prompt template and apply in-context learning to help ChatGPT-4 produce the desired output. The prompt template can be divided into three parts. In the first part, we tell the ChatGPT-4 to play the role of a professional image annotator and specify the requirements of generated questions. In the second part, the template provides an example of the questions and format. In the last part, we construct our queries using the same structure as the example. The image description for $<\text{summary}>$ comes from a LLaVA-v1.5-13B model. The categories for $<\text{important\_objects}>$ come from the ground truth of the segmentation. We keep the questions empty to let ChatGPT-4 fill it. The complete template is shown in Figure \ref{fig:gpt_prompt}.

\subsection{Dataset Analysis}
Our LLM-Seg40K dataset contains 14K images in total. The dataset is divided into training, validation, and test sets, containing 11K, 1K, and 2K images respectively. For the training split, each image has 3.95 questions on average and the average question question length is 15.2 words. The training set contains 1458 different categories in total.

\section{Experiments}
In this section, we present the empirical results for our LLM-Seg model and the LLM-Seg40K dataset.

\subsection{Evaluation of LLM-Seg}
\subsubsection{Implement Details}

\textbf{Model Structure} The vision language part in our method is a pretrained LLaVA-lightning-7B-v1 model. DINOv2-ViT-L is used to extract image features for mask pooling. The feature dimension for the fusion module is set to $256$. For the IoU and IoP selection head, the MLP module has 3 layers with dimensions $[256, 64, 1]$. 

\textbf{Training Datasets} We use a similar training recipe of LISA to train our model. More specifically, the training dataset is a mixture of different semantic segmentation datasets, referring segmentation datasets and the ReasonSeg dataset. Details of the training recipe can be referred to \cite{lai2023lisa}. 

\textbf{Training Setting} The training script is modified from that of LISA \cite{lai2023lisa}. We apply DeepSpeed \cite{rasley2020deepspeed} as the training framework and finish the training with 2 RTX-A6000 GPUs. The training consists of 5000 updating steps and it takes 15 hours to complete. We choose AdamW \cite{loshchilov2017adamW} as the optimizer and apply the WarmupDecayLR to adjust the learning rate. The initial learning rate is set to 0.0001. The batch size for a single GPU is 1 and the gradient accumulation step is set to 10.
For the loss function, we set $\lambda_{iou}$ to 1.0 and $\lambda_{iop}$ to 50.0.

\begin{table}[htbp] 
\centering
\caption{The performance of different methods on ReasonSeg Validation split. ``Use LLM" shows whether a pretrained LLM model is included in the methods. ``Zero-Shot" shows whether the method is fine-tuned using the training split of the ReasonSeg dataset. We mainly compare our LLM-Seg with LISA and report the results of both methods under 10 epochs (default) and 20 epochs training. The results of methods without LLM and LISA under 20 epochs training are cited from \cite{lai2023lisa}.}
\label{tab:reasonseg_res}
\resizebox{1.0\linewidth}{!}{
\begin{tabular}{cccccc} 
\hline
Method              & Use LLM & Zero-Shot & gIoU & cIoU & ncIoU\\ \hline
GRES \cite{liu2023gres}               & $\times$ & \checkmark    & 22.4 & 19.9 &   -   \\
OVSeg  \cite{liang2023ovseg}             & $\times$  & \checkmark     & 28.5 & 18.6 &   -   \\
X-Decoder \cite{zou2023generalized}         & $\times$  & \checkmark     & 22.6 & 17.9 &   -   \\
SEEM  \cite{zou2023segment}               & $\times$  & \checkmark     & 22.5 & 21.2 &   -   \\ \hline
LISA  \cite{lai2023lisa}               & \checkmark   & $\times$    & 51.0 & 50.6 &   48.1   \\
LISA (20 epochs)               & \checkmark   & \checkmark     & 44.4 & 46.0 &   -   \\
LISA (20 epochs)               & \checkmark   & $\times$    & \underline{ 52.9} & \textbf{54.0}  &   -  \\ \hline
LLaVA+GroundingSAM  & \checkmark   & \checkmark     & 47.4 & 34.6 &   34.4    \\
LLM-Seg         & \checkmark      & \checkmark        & 47.4 & 35.4 &  43.5     \\
LLM-Seg          & \checkmark      & $\times$        & 52.3 & \underline{47.5} &      \underline{48.8} \\
LLM-Seg (20 epochs)          & \checkmark      & $\times$        & \textbf{55.4} & 45.6 &   \textbf{51.0}   \\
\hline

\end{tabular}}
\end{table}

\textbf{Evaluation Dataset and Mertics} \label{sec:eval_setting}
We evaluate our model using ReasonSeg \cite{lai2023lisa} validation split. We first use two metrics in \cite{lai2023lisa}: cumulative IoU (cIoU) and generalized IoU (gIoU) \cite{liu2023gres} to evaluate our model. The gIoU is the average IoU of all samples. The cIoU is defined by counting the intersection and union pixels among all the samples and then computing the intersection over the union. However, the computation of the cIoU ignores the image size, we introduce a new metric, the normalized cIoU  (ncIoU), to provide a more comprehensive evaluation of the models. Our proposed ncIoU standardizes the prediction and ground truth to the same size before computation, enabling a fairer comparison between different models.

\subsubsection{Quantitative Results} \label{sec:reasonseg_res}
We evaluate our LLM-Seg on the ReasonSeg validation split and compare the result with other existing methods. For the methods to be compared, we take the results from \cite{lai2023lisa} which includes the existing referring segmentation methods and LISA itself. Besides, we also implement a baseline method by combining a vision language model and an open-vocabulary segmentation model (LLaVA+GroundingSAM). The results of different methods are shown in Table \ref{tab:reasonseg_res}.

The baseline method is implemented based on the LLaVA-v1.5-13B and GroundingSAM. Specifically, we concatenate the original question with the sentence "\texttt{Please answer the question using a word or phrase.}" to prompt the LLaVA model. The answer from LLaVA is then used as the text prompt of GroundingSAM to generate the segmentation.

Our proposed LLM-Seg shows competitive results compared with the current state-of-the-art method LISA. For the gIoU, our LLM-Seg shows a higher gIoU under the same evaluation settings (47.4 vs. 44.4 under the zero-shot setting, 52.3 vs. 51.0 under 10 epochs training, and 55.4 vs. 52.9 under 20 epochs training). However, we also notice the cIoU of our method is about 10 points lower than that of LISA. We argue the low cIoU of LLM-Seg is due to wrong segmentation for high-resolution images within the dataset. Our method is more likely to over-segment targets, therefore the large images will significantly affect the cIoU. To support our argument, we compare the results of ncIoU. We find for LISA, a more balanced method, the ncIoU is 2 points lower than the cIoU. However, the ncIoU of our LLM-Seg is higher than the cIoU, especially under the zero-shot setting where more mistakes exist. Under the same setting, our LLM-Seg has a better performance than LISA in terms of the ncIoU (48.8 vs. 48.1).

\subsubsection{Ablation Study}
\textbf{Model Structure}
To evaluate the effectiveness of different modules in our method, we perform ablations and report the result in Table \ref{tab:ablation_model}. The baseline mode contains the IoU selection head only and we replace the fusion module with an MLP layer and update the mask embedding only. Although adding the fusion module does not help for both metrics, the structure is necessary for adding the IoP head to the model. For the selection heads, we notice that the single-head design cannot give optimal performance. The IoU selection head shows a higher gIoU score while the IoP selection head gives a better cIoU score. Combining both heads gives the optimal performance, even though the final mask selection uses IoP only.

\begin{table}[htbp] 
\centering
\caption{The performance of our method under different model structures. For the model with IoU head only, we pick the mask with the highest similarity. For the model with the IoP head, we pick masks based on a threshold of the predicted IoP and ignore the prediction of the IoU head.}
\label{tab:ablation_model}
\begin{tabular}{ccccc}
\hline
\multicolumn{3}{c}{Model Structure}   & \multirow{2}{*}{gIoU} & \multirow{2}{*}{cIoU} \\ \cline{1-3}
Fusion Module & IoU Head & IoP Head &                       &                       \\ \hline
              & \checkmark       &        & 49.1                  & 31.8                  \\
\checkmark              & \checkmark       &        & 48.6                  & 28.5                  \\
\checkmark              &        & \checkmark       & 51.1                  & 43.6                  \\
\checkmark              & \checkmark       & \checkmark       & \textbf{52.3}                  & \textbf{47.5}                 \\ \hline
\end{tabular}
\end{table}

\textbf{Mask Selection Strategy}
Based on the final double-head model structure, we experiment with different mask selection strategies and the result is shown in Table \ref{tab:ablation_strategy}. Although our model predicts both IoU and IoP for all the mask proposals, we notice that simply setting a threshold for the predicted IoPs can give the optimal performance. 

\begin{table}[htbp] 
\centering
\caption{The performance of our method using different mask selection strategies. Top1 IoP means we only pick the mask with the highest similarity from the IoU head. Threshold IoP means we select all masks that have an IoP prediction above the threshold. Top1 IoU + Threshold IoP means we add the predictions from the above two strategies. Threshold IoP from Top5 IoU means we use the threshold of IoP to pick masks, but only consider masks that have top5 predicted similarities from the IoU head.}
\label{tab:ablation_strategy}
\begin{tabular}{ccc}
\hline
Mask Pick Up Strategy    & gIoU & cIoU \\ \hline
Top1 IoU                 & 49.6 & 35.4 \\
Threshold IoP            & 52.3 & 47.5 \\
Top1 IoU + Threshold IoP & 53.8 & 42.2 \\ 
Threshold IoP from Top5 IoU & 51.4 & 43.6 \\ \hline
\end{tabular}
\end{table}

\textbf{Image Encoder}
 We experiment with different image encoders for generating mask embeddings in our method. The result is reported in Table \ref{tab:ablation_backbone}. We can find that DINOv2 has a clear advantage over SAM and CLIP for both metrics. The gIoU of DINOv2 is 3.0 points higher than that of CLIP and 6.1 points higher than that of SAM. The performance gap emphasizes the importance of semantic information in the extracted image features.

\begin{table}[htbp]
\centering
\caption{The performance of our method using different vision backbone for the mask embedding.}
\label{tab:ablation_backbone}
\begin{tabular}{ccc}
\hline
Vision Backbone & gIoU & cIoU \\ \hline
DINOv2 (ViT-L) \cite{oquab2023dinov2}         & 52.3 & 47.5  \\ 
CLIP (ConvNext-Large) \cite{ilharco_gabriel_2021_5143773}	             & 49.3 & 44.5 \\ 
SAM (ViT-H) \cite{kirillov2023SAM}             & 46.2 & 44.7 \\ \hline
\end{tabular}
\end{table}

\subsubsection{Qualitative Results}
\begin{figure*}[!htbp]
	\centering
	\includegraphics[width=\linewidth]{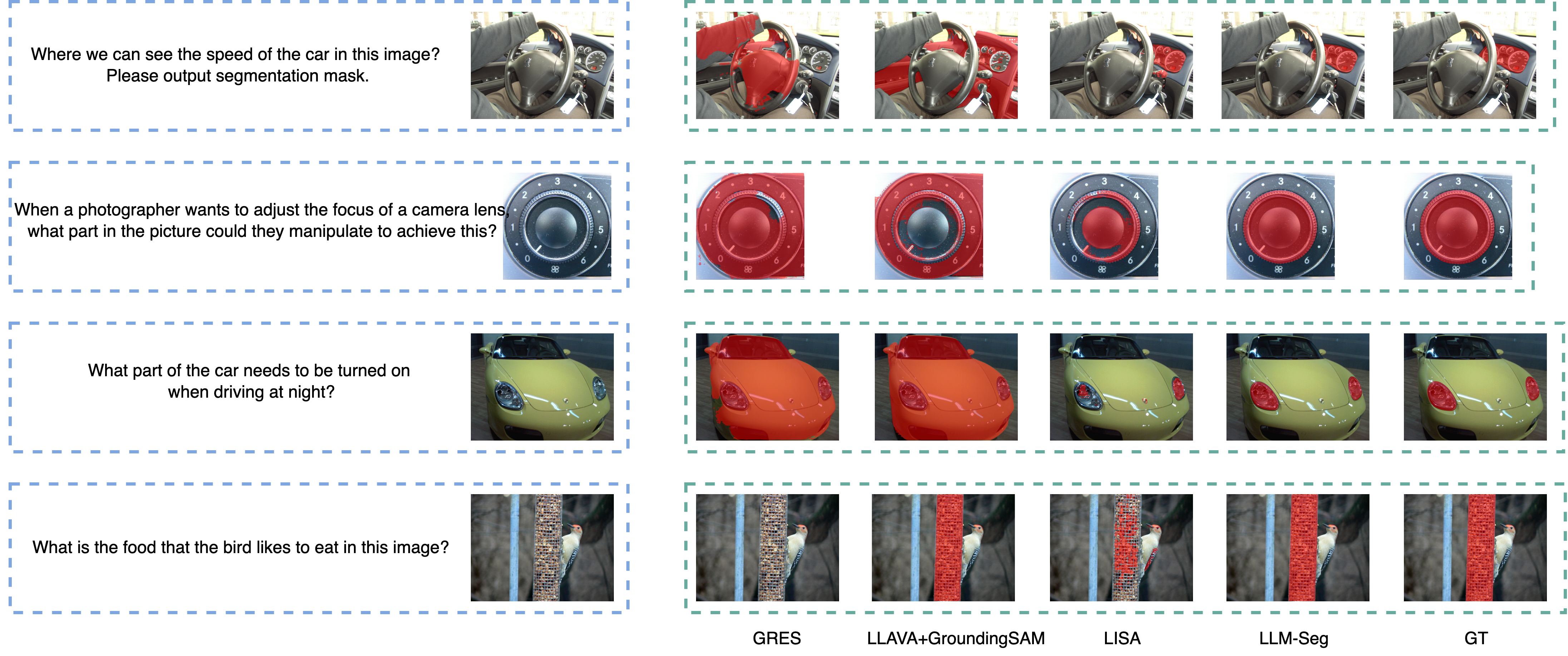}
    \vspace{-0.2in}
	\caption{Visual comparison of LLM-Seg (ours) with the SOTA methods. Our LLM-Seg shows high-quality segmentation results even for multiple instances.}
	\label{fig:qualitative_results}
\end{figure*}

We visualize some cases of the ReasonSeg validation split. Figure \ref{fig:qualitative_results} shows some success cases of our method. Although LISA can give the correct segmentation target, the segmentation quality is lower than that of our LLM-Seg. By comparison, our LLM-Seg keeps the SAM intact and maintains the quality of segmentation masks

We show some failure cases of our LLM-Seg in the supplementary material. One limitation of LLM-Seg is the false positive during mask selection. Because LLM-Seg selects masks with a fixed threshold for the predictions, it is more likely to select a false mask with a large area. The false segmentation area is also consistent with the low cIoU in Table \ref{tab:reasonseg_res}. 

\subsection{Benckmark on LLM-Seg40K dataset}
\begin{figure*}[!htbp]
	\centering
	\includegraphics[width=\linewidth]{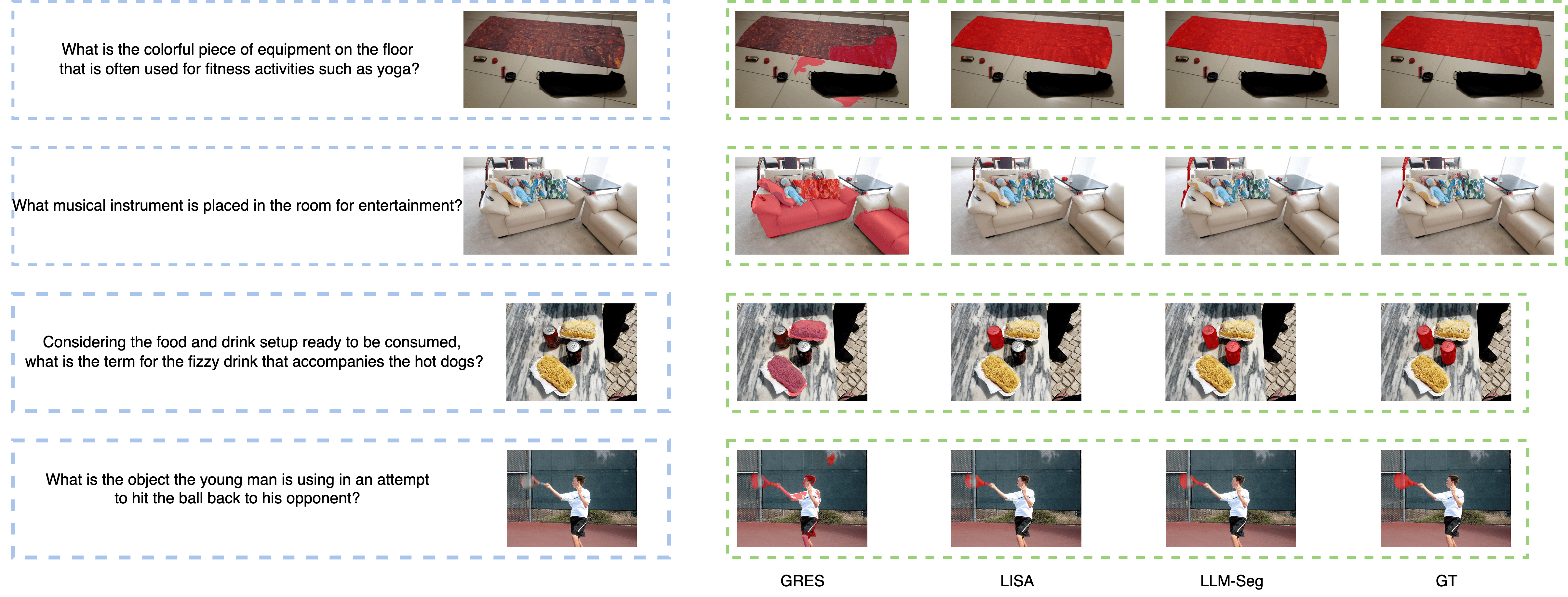}
        \vspace{-0.2in}
	\caption{Visual comparison on our LLM-Seg40K validation split. For LISA and LLM-Seg, results are from the fine-tuned model.}
	\label{fig:llm_seg_res}
	\vspace{-0.1in}
\end{figure*}

We further evaluate different methods using our LLM-Seg40K dataset. The evaluation metrics are similar to Section \ref{sec:eval_setting}, but we remove the ncIoU due to the small variance of image sizes within our dataset. The result is shown in Table \ref{tab:benckmark_res} and we visualize some examples in Figure \ref{fig:llm_seg_res}. 

\subsubsection{Implement Details}
We implement the following methods to evaluate the performance on the LLM-Seg40K validation set:

\textbf{GRES:} We choose GRES~\cite{liu2023gres} as one example of the referring segmentation method. The backbone of the model is Swin-Base.	The evaluation is performed in a zero-shot manner using the released checkpoint. 

\textbf{LISA}: We use the LISA-7B-V1~\cite{lai2023lisa} model for evaluation under two different settings. Under the fully supervised setting, we fine-tune the model using the LLM-Seg40K training split. We maintain most of the training settings of the original LISA. The learning rate is set to $3e^{-5}$ ($1/10$ of the original learning rate), and the batch size is reduced to 20 due to the limited GPU resources. The fine-tuning has 2500 updates in total.

\textbf{LLM-Seg}: For evaluating our proposed LLM-Seg model, we use the checkpoint under 10 epochs training in section \ref{sec:reasonseg_res}. Under the fully supervised setting, the learning rate is set to  $1e^{-5}$. The batch size and number of updates are the same as LISA.

\begin{table}[htbp] 
\centering
\caption{The benchmark of different methods on the validation split of the LLM-Seg40K dataset. ``Use LLM" shows whether a pretrained LLM model is included in the methods. "Zero-Shot" shows whether the method is fine-tuned using the training split of the LLM-Seg40K dataset.}
\label{tab:benckmark_res}
\begin{tabular}{ccccc} 
\hline
Method              & Use LLM & Zero-Shot & gIoU & cIoU \\ \hline
GRES~\cite{liu2023gres}               & $\times$      & \checkmark        & 14.16 & 15.90 \\
LISA~\cite{lai2023lisa}              & \checkmark      & \checkmark        & 33.19 & 37.97 \\
LISA                & \checkmark      & $\times$        & 37.59 & \underline{48.49}  \\ 
LLM-Seg         & \checkmark      & \checkmark        & 36.02 & 39.42 \\
LLM-Seg          & \checkmark      & $\times$        & \textbf{45.47} & \textbf{54.18} \\
\hline

\end{tabular}
\end{table}

\subsubsection{Analysis}
\vspace{-0.1in}
Table \ref{tab:benckmark_res} shows that our LLM-Seg achieves state-of-the-art performance for both metrics under the fully-supervised setting. For the zero-shot setting, our LLM-Seg still shows higher scores than LISA. In addition, we notice that after fine-tuning using the training split of LLM-Seg40K, both LISA and LLM-Seg show performance improvement. However, the improvement of LLM-Seg is much larger than that of LISA. One reason is that unavoidable errors in our dataset bring harm to the training of the mask decoder in LISA. By comparison, the masks of our LLM-Seg are generated by a frozen SAM model. The errors in the training set will not affect the segmentation quality in our method, resulting in a higher performance boost after fine-tuning. This result also verifies the robustness of our LLM-Seg.
\section{Conclusion}

Our work studies the reasoning segmentation from two aspects: the method and the dataset. Our LLM-Seg epitomizes a two-stage method, decoupling the reasoning and segmentation processes to maintain the segmentation quality. Empirical results, both qualitative and quantitative, underscore our LLM-Seg's competitive performance against current state-of-the-art methods. For the dataset part, we propose a novel pipeline and LLM-Seg40K dataset. The experiments show our pipeline can automatically generate high-quality question-segmentation pairs and is useful for model training and validation.

\newpage

{
    \small
    \bibliographystyle{ieeenat_fullname}
    \bibliography{main}

\begin{thebibliography}{40}
\providecommand{\natexlab}[1]{#1}
\providecommand{\url}[1]{\texttt{#1}}
\expandafter\ifx\csname urlstyle\endcsname\relax
  \providecommand{\doi}[1]{doi: #1}\else
  \providecommand{\doi}{doi: \begingroup \urlstyle{rm}\Url}\fi

\bibitem[Alayrac et~al.(2022)Alayrac, Donahue, Luc, Miech, Barr, Hasson, Lenc, Mensch, Millican, Reynolds, et~al.]{alayrac2022flamingo}
Jean-Baptiste Alayrac, Jeff Donahue, Pauline Luc, Antoine Miech, Iain Barr, Yana Hasson, Karel Lenc, Arthur Mensch, Katherine Millican, Malcolm Reynolds, et~al.
\newblock Flamingo: a visual language model for few-shot learning.
\newblock \emph{Advances in Neural Information Processing Systems}, 35:\penalty0 23716--23736, 2022.

\bibitem[Almazrouei et~al.(2023)Almazrouei, Alobeidli, Alshamsi, Cappelli, Cojocaru, Debbah, Goffinet, Hesslow, Launay, Malartic, et~al.]{almazrouei2023falcon}
Ebtesam Almazrouei, Hamza Alobeidli, Abdulaziz Alshamsi, Alessandro Cappelli, Ruxandra Cojocaru, M{\'e}rouane Debbah, {\'E}tienne Goffinet, Daniel Hesslow, Julien Launay, Quentin Malartic, et~al.
\newblock The falcon series of open language models.
\newblock \emph{arXiv preprint arXiv:2311.16867}, 2023.

\bibitem[Bai et~al.(2022)Bai, Kadavath, Kundu, Askell, Kernion, Jones, Chen, Goldie, Mirhoseini, McKinnon, et~al.]{bai2022constitutional}
Yuntao Bai, Saurav Kadavath, Sandipan Kundu, Amanda Askell, Jackson Kernion, Andy Jones, Anna Chen, Anna Goldie, Azalia Mirhoseini, Cameron McKinnon, et~al.
\newblock Constitutional ai: Harmlessness from ai feedback.
\newblock \emph{arXiv preprint arXiv:2212.08073}, 2022.

\bibitem[Brown et~al.(2020)Brown, Mann, Ryder, Subbiah, Kaplan, Dhariwal, Neelakantan, Shyam, Sastry, Askell, et~al.]{brown2020GPT3}
Tom Brown, Benjamin Mann, Nick Ryder, Melanie Subbiah, Jared~D Kaplan, Prafulla Dhariwal, Arvind Neelakantan, Pranav Shyam, Girish Sastry, Amanda Askell, et~al.
\newblock Language models are few-shot learners.
\newblock \emph{Advances in neural information processing systems}, 33:\penalty0 1877--1901, 2020.

\bibitem[Chen et~al.(2023)Chen, Zhang, Zeng, Zhang, Zhu, and Zhao]{chen2023shikra}
Keqin Chen, Zhao Zhang, Weili Zeng, Richong Zhang, Feng Zhu, and Rui Zhao.
\newblock Shikra: Unleashing multimodal llm's referential dialogue magic.
\newblock \emph{arXiv preprint arXiv:2306.15195}, 2023.

\bibitem[Chowdhery et~al.(2023)Chowdhery, Narang, Devlin, Bosma, Mishra, Roberts, Barham, Chung, Sutton, Gehrmann, et~al.]{chowdhery2023palm}
Aakanksha Chowdhery, Sharan Narang, Jacob Devlin, Maarten Bosma, Gaurav Mishra, Adam Roberts, Paul Barham, Hyung~Won Chung, Charles Sutton, Sebastian Gehrmann, et~al.
\newblock Palm: Scaling language modeling with pathways.
\newblock \emph{Journal of Machine Learning Research}, 24\penalty0 (240):\penalty0 1--113, 2023.

\bibitem[Devlin et~al.(2018)Devlin, Chang, Lee, and Toutanova]{devlin2018bert}
Jacob Devlin, Ming-Wei Chang, Kenton Lee, and Kristina Toutanova.
\newblock Bert: Pre-training of deep bidirectional transformers for language understanding.
\newblock \emph{arXiv preprint arXiv:1810.04805}, 2018.

\bibitem[Feng et~al.(2020)Feng, Haase-Sch{\"u}tz, Rosenbaum, Hertlein, Glaeser, Timm, Wiesbeck, and Dietmayer]{feng2020deep_driving}
Di Feng, Christian Haase-Sch{\"u}tz, Lars Rosenbaum, Heinz Hertlein, Claudius Glaeser, Fabian Timm, Werner Wiesbeck, and Klaus Dietmayer.
\newblock Deep multi-modal object detection and semantic segmentation for autonomous driving: Datasets, methods, and challenges.
\newblock \emph{IEEE Transactions on Intelligent Transportation Systems}, 22\penalty0 (3):\penalty0 1341--1360, 2020.

\bibitem[Gupta et~al.(2019)Gupta, Dollar, and Girshick]{gupta2019lvis}
Agrim Gupta, Piotr Dollar, and Ross Girshick.
\newblock Lvis: A dataset for large vocabulary instance segmentation.
\newblock In \emph{Proceedings of the IEEE/CVF conference on computer vision and pattern recognition}, pages 5356--5364, 2019.

\bibitem[Hoffmann et~al.(2022)Hoffmann, Borgeaud, Mensch, Buchatskaya, Cai, Rutherford, Casas, Hendricks, Welbl, Clark, et~al.]{hoffmann2022chinchilla}
Jordan Hoffmann, Sebastian Borgeaud, Arthur Mensch, Elena Buchatskaya, Trevor Cai, Eliza Rutherford, Diego de~Las Casas, Lisa~Anne Hendricks, Johannes Welbl, Aidan Clark, et~al.
\newblock Training compute-optimal large language models.
\newblock \emph{arXiv preprint arXiv:2203.15556}, 2022.

\bibitem[Hu et~al.(2021)Hu, Shen, Wallis, Allen-Zhu, Li, Wang, Wang, and Chen]{hu2021lora}
Edward~J Hu, Yelong Shen, Phillip Wallis, Zeyuan Allen-Zhu, Yuanzhi Li, Shean Wang, Lu Wang, and Weizhu Chen.
\newblock Lora: Low-rank adaptation of large language models.
\newblock \emph{arXiv preprint arXiv:2106.09685}, 2021.

\bibitem[Ilharco et~al.(2021)Ilharco, Wortsman, Wightman, Gordon, Carlini, Taori, Dave, Shankar, Namkoong, Miller, Hajishirzi, Farhadi, and Schmidt]{ilharco_gabriel_2021_5143773}
Gabriel Ilharco, Mitchell Wortsman, Ross Wightman, Cade Gordon, Nicholas Carlini, Rohan Taori, Achal Dave, Vaishaal Shankar, Hongseok Namkoong, John Miller, Hannaneh Hajishirzi, Ali Farhadi, and Ludwig Schmidt.
\newblock Openclip, 2021.
\newblock If you use this software, please cite it as below.

\bibitem[Ke et~al.(2023)Ke, Ye, Danelljan, Liu, Tai, Tang, and Yu]{sam_hq}
Lei Ke, Mingqiao Ye, Martin Danelljan, Yifan Liu, Yu-Wing Tai, Chi-Keung Tang, and Fisher Yu.
\newblock Segment anything in high quality.
\newblock In \emph{NeurIPS}, 2023.

\bibitem[Kirillov et~al.(2023)Kirillov, Mintun, Ravi, Mao, Rolland, Gustafson, Xiao, Whitehead, Berg, Lo, et~al.]{kirillov2023SAM}
Alexander Kirillov, Eric Mintun, Nikhila Ravi, Hanzi Mao, Chloe Rolland, Laura Gustafson, Tete Xiao, Spencer Whitehead, Alexander~C Berg, Wan-Yen Lo, et~al.
\newblock Segment anything.
\newblock \emph{arXiv preprint arXiv:2304.02643}, 2023.

\bibitem[Ko and Lee(2020)]{ko2020novel_ar}
Tae-young Ko and Seung-ho Lee.
\newblock Novel method of semantic segmentation applicable to augmented reality.
\newblock \emph{Sensors}, 20\penalty0 (6):\penalty0 1737, 2020.

\bibitem[Lai et~al.(2023)Lai, Tian, Chen, Li, Yuan, Liu, and Jia]{lai2023lisa}
Xin Lai, Zhuotao Tian, Yukang Chen, Yanwei Li, Yuhui Yuan, Shu Liu, and Jiaya Jia.
\newblock Lisa: Reasoning segmentation via large language model.
\newblock \emph{arXiv preprint arXiv:2308.00692}, 2023.

\bibitem[Li et~al.(2023)Li, Li, Savarese, and Hoi]{li2023blip}
Junnan Li, Dongxu Li, Silvio Savarese, and Steven Hoi.
\newblock Blip-2: Bootstrapping language-image pre-training with frozen image encoders and large language models.
\newblock \emph{arXiv preprint arXiv:2301.12597}, 2023.

\bibitem[Li et~al.(2022)Li, Guo, Shuang, Zhang, and Li]{li2022robotic_vision}
Yong Li, Zhiqiang Guo, Feng Shuang, Man Zhang, and Xiuhua Li.
\newblock Key technologies of machine vision for weeding robots: A review and benchmark.
\newblock \emph{Computers and Electronics in Agriculture}, 196:\penalty0 106880, 2022.

\bibitem[Liang et~al.(2023)Liang, Wu, Dai, Li, Zhao, Zhang, Zhang, Vajda, and Marculescu]{liang2023ovseg}
Feng Liang, Bichen Wu, Xiaoliang Dai, Kunpeng Li, Yinan Zhao, Hang Zhang, Peizhao Zhang, Peter Vajda, and Diana Marculescu.
\newblock Open-vocabulary semantic segmentation with mask-adapted clip.
\newblock In \emph{Proceedings of the IEEE/CVF Conference on Computer Vision and Pattern Recognition}, pages 7061--7070, 2023.

\bibitem[Lin et~al.(2014)Lin, Maire, Belongie, Hays, Perona, Ramanan, Doll{\'a}r, and Zitnick]{lin2014mscoco}
Tsung-Yi Lin, Michael Maire, Serge Belongie, James Hays, Pietro Perona, Deva Ramanan, Piotr Doll{\'a}r, and C~Lawrence Zitnick.
\newblock Microsoft coco: Common objects in context.
\newblock In \emph{Computer Vision--ECCV 2014: 13th European Conference, Zurich, Switzerland, September 6-12, 2014, Proceedings, Part V 13}, pages 740--755. Springer, 2014.

\bibitem[Liu et~al.(2023{\natexlab{a}})Liu, Ding, and Jiang]{liu2023gres}
Chang Liu, Henghui Ding, and Xudong Jiang.
\newblock Gres: Generalized referring expression segmentation.
\newblock In \emph{Proceedings of the IEEE/CVF Conference on Computer Vision and Pattern Recognition}, pages 23592--23601, 2023{\natexlab{a}}.

\bibitem[Liu et~al.(2023{\natexlab{b}})Liu, Li, Li, and Lee]{liu2023improved}
Haotian Liu, Chunyuan Li, Yuheng Li, and Yong~Jae Lee.
\newblock Improved baselines with visual instruction tuning, 2023{\natexlab{b}}.

\bibitem[Liu et~al.(2023{\natexlab{c}})Liu, Li, Wu, and Lee]{liu2023visual}
Haotian Liu, Chunyuan Li, Qingyang Wu, and Yong~Jae Lee.
\newblock Visual instruction tuning.
\newblock \emph{arXiv preprint arXiv:2304.08485}, 2023{\natexlab{c}}.

\bibitem[Loshchilov and Hutter(2017)]{loshchilov2017adamW}
Ilya Loshchilov and Frank Hutter.
\newblock Decoupled weight decay regularization.
\newblock \emph{arXiv preprint arXiv:1711.05101}, 2017.

\bibitem[L{\"u}ddecke and Ecker(2022)]{luddecke2022clipseg}
Timo L{\"u}ddecke and Alexander Ecker.
\newblock Image segmentation using text and image prompts. in 2022 ieee.
\newblock In \emph{CVF Conference on Computer Vision and Pattern Recognition (CVPR)}, pages 7076--7086, 2022.

\bibitem[Neuhold et~al.(2017)Neuhold, Ollmann, Rota~Bulo, and Kontschieder]{neuhold2017mapillary}
Gerhard Neuhold, Tobias Ollmann, Samuel Rota~Bulo, and Peter Kontschieder.
\newblock The mapillary vistas dataset for semantic understanding of street scenes.
\newblock In \emph{Proceedings of the IEEE international conference on computer vision}, pages 4990--4999, 2017.

\bibitem[Oquab et~al.(2023)Oquab, Darcet, Moutakanni, Vo, Szafraniec, Khalidov, Fernandez, Haziza, Massa, El-Nouby, et~al.]{oquab2023dinov2}
Maxime Oquab, Timoth{\'e}e Darcet, Th{\'e}o Moutakanni, Huy Vo, Marc Szafraniec, Vasil Khalidov, Pierre Fernandez, Daniel Haziza, Francisco Massa, Alaaeldin El-Nouby, et~al.
\newblock Dinov2: Learning robust visual features without supervision.
\newblock \emph{arXiv preprint arXiv:2304.07193}, 2023.

\bibitem[Radford et~al.(2021)Radford, Kim, Hallacy, Ramesh, Goh, Agarwal, Sastry, Askell, Mishkin, Clark, et~al.]{radford2021CLIP}
Alec Radford, Jong~Wook Kim, Chris Hallacy, Aditya Ramesh, Gabriel Goh, Sandhini Agarwal, Girish Sastry, Amanda Askell, Pamela Mishkin, Jack Clark, et~al.
\newblock Learning transferable visual models from natural language supervision.
\newblock In \emph{International conference on machine learning}, pages 8748--8763. PMLR, 2021.

\bibitem[Rasley et~al.(2020)Rasley, Rajbhandari, Ruwase, and He]{rasley2020deepspeed}
Jeff Rasley, Samyam Rajbhandari, Olatunji Ruwase, and Yuxiong He.
\newblock Deepspeed: System optimizations enable training deep learning models with over 100 billion parameters.
\newblock In \emph{Proceedings of the 26th ACM SIGKDD International Conference on Knowledge Discovery \& Data Mining}, pages 3505--3506, 2020.

\bibitem[Touvron et~al.(2023)Touvron, Lavril, Izacard, Martinet, Lachaux, Lacroix, Rozi{\`e}re, Goyal, Hambro, Azhar, et~al.]{touvron2023llama}
Hugo Touvron, Thibaut Lavril, Gautier Izacard, Xavier Martinet, Marie-Anne Lachaux, Timoth{\'e}e Lacroix, Baptiste Rozi{\`e}re, Naman Goyal, Eric Hambro, Faisal Azhar, et~al.
\newblock Llama: Open and efficient foundation language models.
\newblock \emph{arXiv preprint arXiv:2302.13971}, 2023.

\bibitem[Wang et~al.(2023)Wang, Lv, Yu, Hong, Qi, Wang, Ji, Yang, Zhao, Song, et~al.]{wang2023cogvlm}
Weihan Wang, Qingsong Lv, Wenmeng Yu, Wenyi Hong, Ji Qi, Yan Wang, Junhui Ji, Zhuoyi Yang, Lei Zhao, Xixuan Song, et~al.
\newblock Cogvlm: Visual expert for pretrained language models.
\newblock \emph{arXiv preprint arXiv:2311.03079}, 2023.

\bibitem[Wang et~al.(2022)Wang, Lu, Li, Tao, Guo, Gong, and Liu]{wang2022cris}
Zhaoqing Wang, Yu Lu, Qiang Li, Xunqiang Tao, Yandong Guo, Mingming Gong, and Tongliang Liu.
\newblock Cris: Clip-driven referring image segmentation.
\newblock In \emph{Proceedings of the IEEE/CVF conference on computer vision and pattern recognition}, pages 11686--11695, 2022.

\bibitem[Xiao et~al.(2023)Xiao, Wu, Xu, Dai, Hu, Lu, Zeng, Liu, and Yuan]{xiao2023florence2}
Bin Xiao, Haiping Wu, Weijian Xu, Xiyang Dai, Houdong Hu, Yumao Lu, Michael Zeng, Ce Liu, and Lu Yuan.
\newblock Florence-2: Advancing a unified representation for a variety of vision tasks.
\newblock \emph{arXiv preprint arXiv:2311.06242}, 2023.

\bibitem[Yu et~al.(2023)Yu, He, Deng, Shen, and Chen]{yu2023fcclip}
Qihang Yu, Ju He, Xueqing Deng, Xiaohui Shen, and Liang-Chieh Chen.
\newblock Convolutions die hard: Open-vocabulary segmentation with single frozen convolutional clip.
\newblock \emph{arXiv preprint arXiv:2308.02487}, 2023.

\bibitem[Yuan et~al.(2021)Yuan, Chen, Chen, Codella, Dai, Gao, Hu, Huang, Li, Li, et~al.]{yuan2021florence}
Lu Yuan, Dongdong Chen, Yi-Ling Chen, Noel Codella, Xiyang Dai, Jianfeng Gao, Houdong Hu, Xuedong Huang, Boxin Li, Chunyuan Li, et~al.
\newblock Florence: A new foundation model for computer vision.
\newblock \emph{arXiv preprint arXiv:2111.11432}, 2021.

\bibitem[Zhang et~al.(2023)Zhang, Han, Qiao, Kim, Bae, Lee, and Hong]{mobile_sam}
Chaoning Zhang, Dongshen Han, Yu Qiao, Jung~Uk Kim, Sung-Ho Bae, Seungkyu Lee, and Choong~Seon Hong.
\newblock Faster segment anything: Towards lightweight sam for mobile applications.
\newblock \emph{arXiv preprint arXiv:2306.14289}, 2023.

\bibitem[Zhou et~al.(2019)Zhou, Zhao, Puig, Xiao, Fidler, Barriuso, and Torralba]{zhou2019ade20k}
Bolei Zhou, Hang Zhao, Xavier Puig, Tete Xiao, Sanja Fidler, Adela Barriuso, and Antonio Torralba.
\newblock Semantic understanding of scenes through the ade20k dataset.
\newblock \emph{International Journal of Computer Vision}, 127:\penalty0 302--321, 2019.

\bibitem[Zhu et~al.(2023)Zhu, Xiao, Alvarado, Babaei, Hu, El-Mohri, Culatana, Sumbaly, and Yan]{zhu2023egoobjects}
Chenchen Zhu, Fanyi Xiao, Andr{\'e}s Alvarado, Yasmine Babaei, Jiabo Hu, Hichem El-Mohri, Sean Culatana, Roshan Sumbaly, and Zhicheng Yan.
\newblock Egoobjects: A large-scale egocentric dataset for fine-grained object understanding.
\newblock In \emph{Proceedings of the IEEE/CVF International Conference on Computer Vision}, pages 20110--20120, 2023.

\bibitem[Zou et~al.(2023{\natexlab{a}})Zou, Dou, Yang, Gan, Li, Li, Dai, Behl, Wang, Yuan, et~al.]{zou2023generalized}
Xueyan Zou, Zi-Yi Dou, Jianwei Yang, Zhe Gan, Linjie Li, Chunyuan Li, Xiyang Dai, Harkirat Behl, Jianfeng Wang, Lu Yuan, et~al.
\newblock Generalized decoding for pixel, image, and language.
\newblock In \emph{Proceedings of the IEEE/CVF Conference on Computer Vision and Pattern Recognition}, pages 15116--15127, 2023{\natexlab{a}}.

\bibitem[Zou et~al.(2023{\natexlab{b}})Zou, Yang, Zhang, Li, Li, Gao, and Lee]{zou2023segment}
Xueyan Zou, Jianwei Yang, Hao Zhang, Feng Li, Linjie Li, Jianfeng Gao, and Yong~Jae Lee.
\newblock Segment everything everywhere all at once.
\newblock \emph{arXiv preprint arXiv:2304.06718}, 2023{\natexlab{b}}.

\end{thebibliography}
}

\end{document}